\documentclass[letterpaper, 10 pt, conference]{ieeeconf}  % Comment this line out if you need a4paper

\IEEEoverridecommandlockouts                              % This command is only needed if
\usepackage{graphicx}
\usepackage{cite}
\usepackage[colorlinks=true,
            citecolor=blue,
            linkcolor=blue,
            urlcolor=blue]{hyperref}
\usepackage{amsmath}
\usepackage{amssymb}
\usepackage{booktabs}
\usepackage{multirow}
\usepackage{xcolor}
\usepackage{tikz}
\usetikzlibrary{arrows.meta,positioning,fit,backgrounds,calc}

\hypersetup{
  colorlinks=true,
  citecolor=blue,
  linkcolor=blue,
  urlcolor=blue,
  linktoc=all
}

\title{\LARGE \bf
Imagining the Sense of Touch: Touch-Informed Manipulation via Imagined Tactile Representations
}

\author{
Zhiyuan~Zhang$^{*}$, 
Adeesh~Desai$^{*}$, 
Jyun-Chi~Hu,
Yosuke~Saka,
Quan~Khanh~Luu,
Jiuzhou~Lei,\\
Davood~Soleymanzadeh,
Bihao~Zhang,
Minghui~Zheng, 
Yu~She$^{\dagger}$%
\thanks{This work was partially supported by the United States Department of Agriculture (USDA) under Grant Nos. 2023-67021-39072 and 2024-67021-42878, and by the National Science Foundation (NSF) under Grant Nos. 2423068 and 2520136.}
\thanks{$^{*}$These authors contributed equally. 
$^{\dagger}$Corresponding author.}%
\thanks{Zhiyuan, Adeesh, Jyun-Chi, Yosuke, Quan, and Yu are with the Department of Industrial Engineering, Purdue University, West Lafayette, IN 47907, USA 
{\tt\footnotesize \{zhan5570, desai274, hu1148, ysaka, luu15, shey\}@purdue.edu}}%
\thanks{Jiuzhou, Davood, Bihao, and Minghui are with the Department of Mechanical Engineering, Texas A\&M University, College Station, TX 77843, USA 
{\tt\footnotesize \{jiuzl, davoodso, bhzhang, mhzheng\}@tamu.edu}}%
}

\begin{document}

\maketitle
\thispagestyle{empty}
\pagestyle{empty}

%%%%%%%%%%%%%%%%%%%%%%%%%%%%%%%%%%%%%%%%%%%%%%%%%%%%%%%%%%%%%%%%%%%%%%%%%%%%%%%%
\begin{abstract}
Tactile sensing can substantially improve contact-rich robotic manipulation, yet its practical deployment remains limited by the fragility, calibration requirements, and maintenance burden of tactile hardware. This raises a fundamental question: \emph{can robots benefit from tactile knowledge without requiring tactile sensors at deployment?}
We present \textbf{TacImag}, a tactile imagination framework that predicts tactile observations from vision and proprioception and uses the generated signals to guide manipulation policies. Trained from paired visuotactile demonstrations, TacImag enables touch-informed manipulation using only visual observations at test time.
We evaluate TacImag on six simulated and four real-world manipulation tasks. Across simulation and real-world experiments, imagined tactile observations consistently improve manipulation performance without requiring tactile hardware. In real-world experiments, imagined force fields improve contact-sensitive tasks by 44.4\% on average, whereas imagined tactile images improve texture-sensitive tasks by 23.3\%, revealing that the effectiveness of tactile imagination depends strongly on the relationship between tactile representation and task requirements.
Our results further suggest that tactile imagination does not simply recover missing tactile measurements. Instead, it acts as a form of contact-aware supervision that transforms subtle visual interaction cues into representations that are easier for manipulation policies to exploit.
\textbf{Project Website:} \url{https://tacimag.github.io/}
\end{abstract}

%%%%%%%%%%%%%%%%%%%%%%%%%%%%%%%%%%%%%%%%%%%%%%%%%%%%%%%%%%%%%%%%%%%%%%%%%%%%%%%%
\section{INTRODUCTION}
\begin{figure*}[t]
    \centering
    \includegraphics[width=\textwidth]{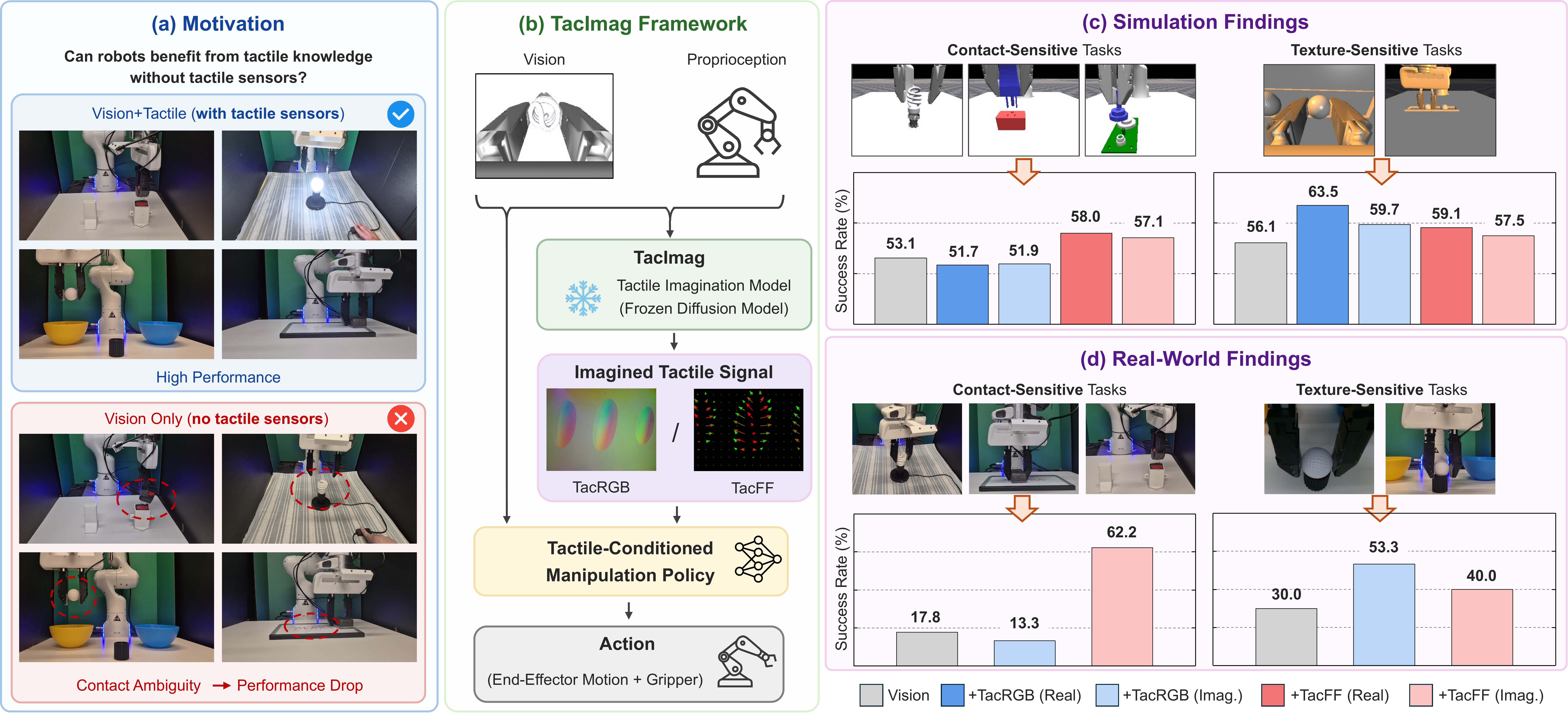}
    \caption{
    \textbf{TacImag: Touch-informed manipulation without tactile sensors.}
    (\textbf{a}) Motivation. Vision-only policies often fail in contact-rich manipulation because contact states are difficult to infer from visual observations alone. In contrast, policies equipped with tactile sensing can directly access contact information and achieve substantially higher performance.
    (\textbf{b}) TacImag framework. TacImag first trains a tactile imagination model to predict tactile observations from visual observations and proprioceptive states. The trained imagination model is then frozen and used to generate imagined tactile signals, which are subsequently consumed by a tactile-conditioned manipulation policy.
    (\textbf{c}) Simulation findings. The effectiveness of tactile imagination depends strongly on tactile representation and task characteristics. For contact-sensitive tasks, imagined TacFF closely matches the performance of real tactile sensing and consistently outperforms imagined TacRGB. For texture-sensitive tasks, imagined TacRGB provides the largest performance gains.
    (\textbf{d}) Real-world findings. Similar trends are observed on real robotic manipulation tasks. Imagined TacFF substantially improves performance on contact-sensitive tasks, while imagined TacRGB is most effective for texture-sensitive object discrimination.
    }
    \label{fig:overview}
\end{figure*}
Contact-rich manipulation, including insertion, assembly, screwing, and sorting, is a fundamental capability for robots operating in unstructured environments.
Recent advances in imitation learning have enabled impressive progress on such tasks using visual observations alone~\cite{chi2023diffusion,zhao2023aloha,ze20243d,zhang2025canonical}.
However, purely visual policies often struggle precisely when contact becomes critical.
Under tight tolerances, occlusion, ambiguous appearance, or degraded lighting, the forces and micro-contacts that determine task success are difficult or impossible to infer directly from camera observations~\cite{luu2025manifeel,calandra2018more}.
Tactile sensing provides complementary information about physical interaction and has repeatedly been shown to improve robustness in contact-rich manipulation~\cite{luu2025manifeel,xue2025reactive,huang20243dvitac,saka2026contact,zhang2026contactworld}.
High-resolution tactile sensors such as GelSight~\cite{yuan2017gelsight,wang2021gelsight} and their simulated counterparts~\cite{akinola2025tacsl} have therefore become increasingly important components of modern robotic manipulation systems.

Despite their effectiveness, tactile sensors remain difficult to deploy in practice.
They require specialized hardware, careful calibration, additional wiring and synchronization, and are often susceptible to wear and damage during prolonged contact.
These challenges motivate a practical question: \emph{can robots benefit from tactile knowledge without requiring tactile sensors at deployment time?}
Such a capability would allow robots to leverage rich visuotactile training data while retaining the simplicity and reliability of vision-only hardware during execution.

We address this question through \textbf{Tactile Imagination (TacImag)} (Fig.~\ref{fig:overview}).
Given paired visual and tactile demonstrations, TacImag learns a tactile imagination model that predicts tactile observations from visual observations and proprioceptive states.
After training, the imagination model is frozen and used to generate tactile observations online, which are subsequently provided as additional inputs to a tactile-conditioned manipulation policy.
At deployment, the robot operates entirely from visual observations while internally generating imagined tactile signals, eliminating the need for physical tactile sensors.
Fig.~\ref{fig:overview} provides an overview of the motivation, framework, and key findings of TacImag.

A natural intuition is that imagined touch could compensate for information missing from vision.
However, in many manipulation scenarios, contact-relevant information is not entirely absent from visual observations.
Instead, it may be weakly observable through subtle appearance changes, object deformation, relative motion, or contact-induced visual cues that are difficult to exploit through direct end-to-end policy learning.
We therefore hypothesize that tactile imagination can serve as a form of contact-aware supervision, encouraging policies to identify and leverage visual features associated with physical interaction.
Under this view, the benefit of tactile imagination arises not necessarily from recovering unseen information, but from exposing contact-relevant structure already latent in vision.

To investigate this hypothesis, we instantiate TacImag using two tactile representations with distinct characteristics:
high-dimensional tactile RGB images (TacRGB) and compact tactile force fields (TacFF).
We evaluate TacImag across six contact-rich manipulation tasks in simulation and five real-world manipulation tasks.
Across these tasks, imagined tactile observations provide a useful source of contact-related information for manipulation policies and can, in many cases, recover a substantial portion of the benefit provided by real tactile sensing.

A consistent pattern emerges across both simulation and real-world experiments, as summarized in Fig.~\ref{fig:overview}(c,d), the effectiveness of tactile imagination depends strongly on both tactile representation and task characteristics.
Imagined TacFF is most effective for contact-sensitive tasks such as insertion, assembly, and screwing, where force and contact-state information are critical for success.
In contrast, imagined TacRGB provides larger gains in tasks that rely on surface appearance or texture discrimination.
Furthermore, we find that tactile imagination is most beneficial when contact-relevant information is already partially encoded in visual observations. These findings support our hypothesis that tactile imagination acts primarily as a contact-aware supervisory signal, helping policies exploit subtle visual cues associated with physical interaction while substantially reducing dependence on tactile hardware.

This paper makes the following contributions:
\begin{enumerate}
\item We introduce \textbf{TacImag}, a tactile imagination framework that couples a frozen vision-to-touch imagination model with a tactile-conditioned manipulation policy, enabling touch-informed control without requiring tactile hardware at deployment.
\item We systematically investigate tactile imagination across two tactile representations, TacRGB and TacFF, revealing that the effectiveness of imagined tactile information depends strongly on both tactile representation and task characteristics.
\item We provide empirical evidence that tactile imagination does not hallucinate information absent from vision. Instead, TacImag transforms visually observable but difficult-to-exploit interaction cues into tactile representations that are easier for manipulation policies to utilize, offering new insight into how tactile knowledge can be recovered from visual observations.
\end{enumerate}

%%%%%%%%%%%%%%%%%%%%%%%%%%%%%%%%%%%%%%%%%%%%%%%%%%%%%%%%%%%%%%%%%%%%%%%%%%%%%%%%
\section{RELATED WORK}
\subsection{Visuotactile Manipulation}
Tactile sensing provides rich information about contact geometry, force, and interaction dynamics that is often unavailable from vision alone.
Vision-based tactile sensors such as GelSight~\cite{yuan2017gelsight,wang2021gelsight} and their simulated counterparts in TacSL~\cite{akinola2025tacsl} have enabled learning-based manipulation systems that jointly leverage vision and touch~\cite{xue2025reactive,huang20243dvitac,heng2025vitacformer}.
Recent benchmarks and studies, including ManiFeel~\cite{luu2025manifeel} and ContactWorld~\cite{zhang2026contactworld}, have systematically investigated the role of tactile sensing in contact-rich manipulation.
A consistent observation across these works is that tactile representation plays a critical role in downstream performance: compact force-based representations often provide more useful control signals than high-dimensional tactile imagery for insertion, assembly, and screwing tasks.

Parallel efforts have focused on learning transferable tactile representations through large-scale pretraining and self-supervised learning, including UniT~\cite{xu2025unit}, T3~\cite{zhao2024t3}, AnyTouch~\cite{feng2025anytouch}, and Sparsh~\cite{higuera2024sparsh}.
However, most existing visuotactile manipulation approaches rely on physical tactile sensors during both training and deployment. In contrast, our goal is to retain the benefits of tactile information while eliminating tactile hardware during execution.

\subsection{Tactile Imagination and Missing-Modality Learning}
A growing body of work explores cross-modal generation between vision and touch.
Early studies demonstrated that tactile observations can be predicted from visual inputs and vice versa, revealing strong correlations between the two sensing modalities~\cite{li2019connecting}.
More recently, diffusion-based approaches have enabled increasingly realistic tactile synthesis.
Touch2Touch~\cite{rodriguez2024touch2touch} and Cross-Sensor Touch Generation~\cite{rodriguez2025crosssensor} translate observations between different tactile sensors, while TactileGen~\cite{lin2024tactilegen} synthesizes tactile observations from visual inputs and contact conditions for simulation and sim-to-real transfer.
Imagine2Touch~\cite{ayad2024imagine2touch} similarly predicts tactile signals from visual observations to anticipate physical interactions. These works demonstrate that tactile observations can be inferred from other sensing modalities, but primarily evaluate generation fidelity, perception accuracy, or sensor translation performance rather than downstream manipulation.

Our work is also related to learning paradigms that exploit privileged modalities during training while operating with reduced sensing at deployment.
Prior work has explored privileged learning and policy distillation~\cite{chen2020learningbycheating}, as well as modality hallucination networks that infer missing modalities as auxiliary supervision~\cite{hoffman2016hallucination}.
In robotic manipulation, privileged tactile latent distillation transfers tactile information into deployable policies through latent-space supervision~\cite{chen2026ptld}, while tactile-conditioned diffusion policies directly utilize tactile observations during execution~\cite{helmut2025tactilecond}.
These approaches demonstrate that information from unavailable modalities can be transferred into deployable policies, but typically encode the privileged information implicitly within the policy parameters.
As a result, it is often difficult to analyze how the missing modality contributes to downstream decision making.

TacImag combines these two perspectives: it uses cross-modal generation to predict tactile observations while addressing the missing-modality setting where tactile sensing is unavailable during deployment.
Unlike prior work that focuses on generation quality, perception, or latent distillation, TacImag investigates how imagined tactile observations influence downstream manipulation performance.

\section{TACTILE IMAGINATION FRAMEWORK}
\begin{figure*}[t]
    \centering
    \includegraphics[width=0.8\textwidth]{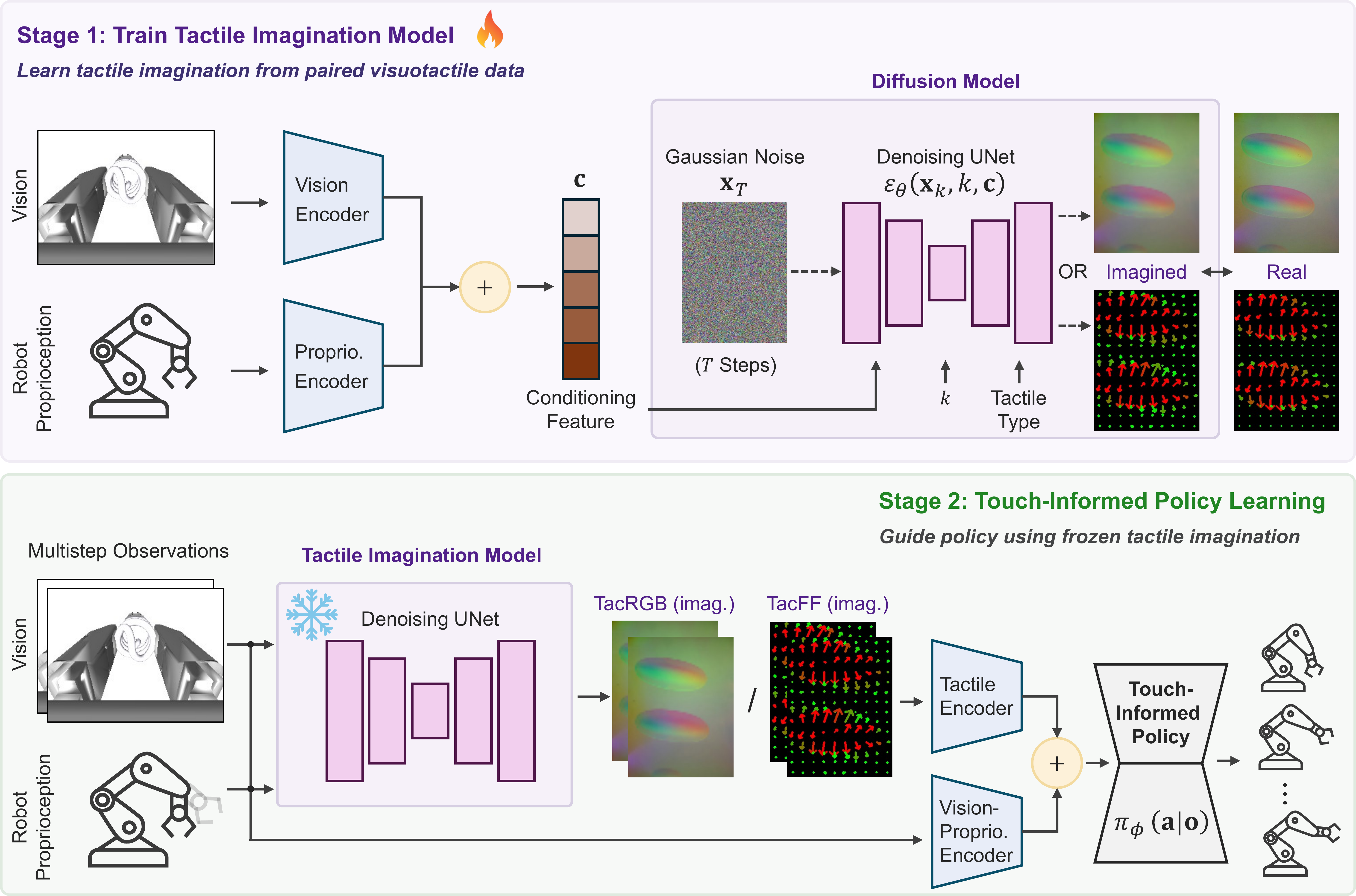}
    \caption{
    \textbf{TacImag framework.}
    TacImag enables touch-informed manipulation without tactile sensors at deployment.
    In Stage 1, a tactile imagination model is learned from paired visuotactile demonstrations using a conditional diffusion model that predicts tactile observations from visual observations and proprioceptive states.
    Separate models are trained for TacRGB and TacFF representations.
    After training, the imagination model is frozen.
    In Stage 2, imagined tactile observations generated by the frozen model are combined with visual and proprioceptive information to train a touch-informed manipulation policy.
    At deployment, the robot operates using only visual observations and proprioceptive states while internally generating tactile representations online.
    }
    \label{fig:framework}
\end{figure*}
\subsection{Problem Setup}
We consider contact-rich manipulation as a supervised imitation learning problem.
A demonstration dataset
\(
\mathcal{D}=\{(\mathbf{o}_t,\mathbf{a}_t)\}_{t=1}^{N}
\)
consists of multimodal observations and expert actions.
At each timestep \(t\), the observation is
\begin{equation}
\mathbf{o}_t=
\{
\mathbf{o}^{\mathrm{vis}}_t,
\mathbf{o}^{\mathrm{tac}}_t,
\mathbf{o}^{\mathrm{prop}}_t
\},
\end{equation}
where
\(\mathbf{o}^{\mathrm{vis}}_t\) denotes visual observations,
\(\mathbf{o}^{\mathrm{tac}}_t\) denotes tactile observations,
and
\(\mathbf{o}^{\mathrm{prop}}_t\) denotes proprioceptive robot states.
Following prior visuotactile benchmarks~\cite{luu2025manifeel}, we consider two tactile representations:
tactile RGB images (TacRGB) and tactile force fields (TacFF).
The action
\(\mathbf{a}_t\)
corresponds to a relative end-effector motion command together with an optional gripper action.

Conventional visuotactile policies require tactile observations during both training and deployment,
\begin{equation}
\pi
(
\mathbf{a}_t
|
\mathbf{o}^{\mathrm{vis}}_t,
\mathbf{o}^{\mathrm{tac}}_t,
\mathbf{o}^{\mathrm{prop}}_t
),
\end{equation}
which necessitates physical tactile sensors at execution time.
Our goal is to preserve the benefits of tactile information while eliminating tactile hardware during deployment.

To this end, we introduce \textbf{TacImag}, a tactile imagination framework consisting of two components:
(1) a tactile imagination model that infers tactile observations from visual and proprioceptive inputs, and
(2) a tactile-conditioned manipulation policy that consumes the imagined tactile observations, as illustrated in Fig.~\ref{fig:framework}.
The imagination model is defined as
\begin{equation}
G_\phi:
(
\mathbf{o}^{\mathrm{vis}}_t,
\mathbf{o}^{\mathrm{prop}}_t
)
\rightarrow
\hat{\mathbf{o}}^{\mathrm{tac}}_t,
\end{equation}
where
\(\hat{\mathbf{o}}^{\mathrm{tac}}_t\)
denotes an imagined tactile observation.
The generated tactile signal is subsequently used in place of real tactile measurements during policy learning and deployment.

\subsection{Tactile Imagination}
The objective of tactile imagination is to generate plausible tactile observations conditioned on visual observations and proprioceptive robot states.
Predicting tactile observations from vision is inherently ambiguous, since multiple contact states may correspond to similar visual appearances under occlusion, limited viewpoints, and partial observability. Rather than producing a deterministic prediction, TacImag models a conditional distribution over tactile observations,
\begin{equation}
p
\left(
\mathbf{o}^{\mathrm{tac}}_t
\mid
\mathbf{o}^{\mathrm{vis}}_t,
\mathbf{o}^{\mathrm{prop}}_t
\right).
\end{equation}

To instantiate this idea, we employ a conditional denoising diffusion probabilistic model (DDPM)~\cite{ho2020denoising}. Let \(\mathbf{x}^{\mathrm{tac}}_0\) denote a tactile observation sampled from the dataset. During training, Gaussian noise is progressively added to obtain a noisy tactile sample \(\mathbf{x}^{\mathrm{tac}}_k\) at diffusion step \(k\). The model then learns to predict the injected noise conditioned on visual observations and robot states using the standard DDPM objective,
\begin{equation}
\mathcal{L}_{\mathrm{diff}}
=
\mathbb{E}_{\mathbf{x}^{\mathrm{tac}}_0,k,\boldsymbol{\epsilon}}
\left[
\left\|
\boldsymbol{\epsilon}
-
\boldsymbol{\epsilon}_{\phi}
\left(
\mathbf{x}^{\mathrm{tac}}_k,
k,
\mathbf{o}^{\mathrm{vis}}_t,
\mathbf{o}^{\mathrm{prop}}_t
\right)
\right\|_2^2
\right].
\end{equation}
where $\boldsymbol{\epsilon}\sim\mathcal{N}(0,\mathbf{I})$ is the injected Gaussian noise,
and $\boldsymbol{\epsilon}_{\phi}$ is the diffusion network parameterized by $\phi$.

During inference, tactile observations are generated through iterative denoising conditioned on the visual and proprioceptive inputs, yielding an imagined tactile observation
\begin{equation}
\label{eq:tacimag}
\hat{\mathbf{o}}^{\mathrm{tac}}_t=
G_\phi
\left(
\mathbf{o}^{\mathrm{vis}}_t,
\mathbf{o}^{\mathrm{prop}}_t
\right),
\end{equation}
where \(G_\phi\) denotes the learned tactile imagination model.

Importantly, the objective of TacImag is not to perfectly reconstruct tactile observations. Instead, we seek to generate tactile representations that preserve manipulation-relevant contact information. Consequently, the quality of tactile imagination is evaluated through downstream manipulation performance rather than reconstruction fidelity alone.

We instantiate the imagination model for two tactile representations with different structural properties. 
TacRGB preserves detailed contact appearance and surface texture, whereas TacFF encodes distributed normal and shear forces over a spatial grid.
Training separate imagination models for TacRGB and TacFF enables a controlled study of tactile imagination across different tactile representations and allows us to investigate how the type of imagined tactile information influences downstream manipulation performance.

\subsection{Policy Learning with Imagined Tactile}
As shown in Stage 2 of Fig.~\ref{fig:framework}, the learned tactile imagination model is frozen and used to generate imagined tactile observations for policy learning.
To evaluate the utility of imagined tactile observations for manipulation, we employ a tactile-conditioned imitation learning policy based on Diffusion Policy~\cite{chi2023diffusion}.

The key distinction between TacImag and conventional visuotactile policies is that tactile information is synthesized rather than measured.
Instead of relying on physical tactile sensors, TacImag supplies imagined tactile observations generated by the frozen imagination model in Eq.~\ref{eq:tacimag}.

Each modality is encoded independently and fused into a shared observation representation,
\begin{equation}
\mathbf{z}_t=\{
\mathbf{z}^{\mathrm{vis}}_t
\oplus
\mathbf{z}^{\mathrm{tac}}_t
\oplus
\mathbf{z}^{\mathrm{prop}}_t\},
\end{equation}
where $\oplus$ denotes feature concatenation. The imagined tactile observations are processed using the same tactile encoder architecture employed by policies trained with real tactile sensing, enabling a direct comparison between real and imagined tactile information.

Following the two-stage training procedure shown in Fig.~\ref{fig:framework}, gradients are not propagated into the imagination model during policy training.
As a result, the tactile generator remains fixed and serves as an explicit source of imagined tactile information rather than being implicitly absorbed into the policy parameters.
This design isolates the effect of tactile imagination and allows us to systematically study how different imagined tactile representations influence manipulation performance.

At deployment, the robot operates entirely without tactile hardware.
Visual observations and proprioceptive states are passed through the frozen imagination model to generate tactile representations online, enabling touch-informed manipulation using only vision and proprioceptive sensing.

%%%%%%%%%%%%%%%%%%%%%%%%%%%%%%%%%%%%%%%%%%%%%%%%%%%%%%%%%%%%%%%%%%%%%%%%%%%%%%%%
\section{Simulation Experiments}
\begin{figure*}[t]
    \centering
    \includegraphics[width=\textwidth]{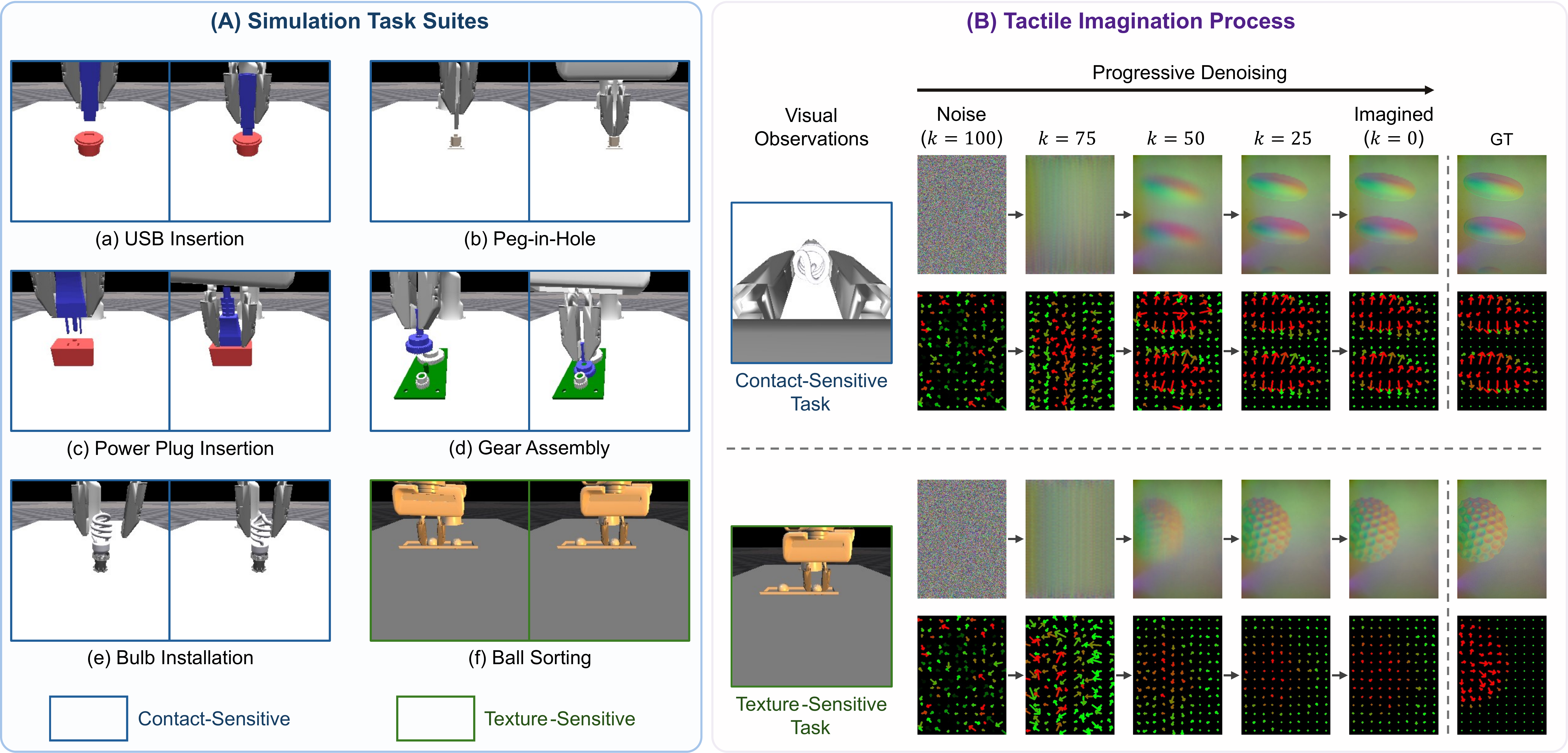}
    \caption{
    \textbf{Simulation tasks and tactile imagination process.}
    \textbf{(A)} Simulation benchmark used for evaluating TacImag.
    Tasks (a)--(e) are contact-sensitive manipulation tasks, including insertion and assembly scenarios where task success primarily depends on accurate contact interactions.
    Task (f) is a texture-sensitive sorting task that relies on object appearance and surface texture.
    For each task, the left image shows the initial state and the right image shows the goal state.
    Contact-sensitive tasks are trained and evaluated using wrist-view observations, whereas the texture-sensitive sorting task uses front-view observations.
    \textbf{(B)} Examples of tactile imagination using conditional diffusion models.
    Given visual observations and proprioceptive states, the model progressively denoises a tactile signal from Gaussian noise.
    Results are shown for representative contact-sensitive (top) and texture-sensitive (bottom) tasks using both TacRGB and TacFF representations.
    From left to right, the tactile representation evolves from Gaussian noise ($k=100$) to the final imagined tactile observation ($k=0$), which closely matches the corresponding ground-truth tactile signal.
    }
    \label{fig:sim_tasks}
\end{figure*}
\subsection{Simulation Setup}
\textbf{Tasks.}
We evaluate TacImag on six contact-rich manipulation tasks in the ManiFeel benchmark~\cite{luu2025manifeel} built on TacSL~\cite{akinola2025tacsl} and IsaacGym~\cite{makoviychuk2021isaac}: USB insertion, power-plug insertion, peg-in-hole (PIH), gear assembly, bulb installation, and ball sorting under dim lighting.
As illustrated in Fig.~\ref{fig:sim_tasks}(A), the benchmark includes five contact-sensitive tasks (USB, Power, PIH, Gear, and Bulb) and one texture-sensitive task (Sorting), covering tight-tolerance insertion, multi-stage assembly, torque-sensitive manipulation, and visually degraded perception scenarios.

\textbf{Observations and tactile representations.}
For the five contact-sensitive tasks (USB, Power, PIH, Gear, and Bulb), the agent receives wrist-view RGB observations.
The wrist camera provides a close-up view of the interaction region but becomes heavily occluded during contact-rich manipulation, creating scenarios where tactile feedback is critical for precise alignment and insertion.
In contrast, the texture-sensitive sorting task uses a front-view camera under dim lighting.
Unlike a wrist-mounted camera, the front view does not reveal fine-grained surface details of the objects, preventing the policy from relying solely on visual appearance for object classification and encouraging the use of tactile information for discrimination.
TacSL provides both a GelSight-style tactile image (TacRGB) and a tactile force-field representation (TacFF)~\cite{wang2021gelsight}.
Object poses and insertion targets are randomized at the beginning of each episode.

\textbf{Sensing configurations.}
For each task, we train five policy variants:
(i) \emph{Vision}, which uses only visual observations and proprioceptive states;
(ii) \emph{Vision+TacRGB} and (iii) \emph{Vision+TacFF}, which additionally receive physical tactile RGB images and tactile force fields, respectively; and
(iv) \emph{Vision+TacRGB (imag.)} and (v) \emph{Vision+TacFF (imag.)}, which replace physical tactile measurements with imagined tactile observations generated by a frozen TacImag model.
The latter two settings require no tactile sensor during deployment.
The imagined variants evaluate whether tactile sensing can be replaced by tactile imagination without requiring physical tactile hardware during deployment.
All policy variants share the same policy architecture, visual encoder, and training procedure, enabling controlled comparisons between tactile representations (TacRGB vs.\ TacFF) and tactile sources (physical vs.\ imagined).

\textbf{Training and evaluation.}
Visual observations are resized to $256\times256\times3$, TacRGB observations have a resolution of $240\times320\times3$, and TacFF observations are represented as a $10\times14\times3$ force-field grid.
We use observation, action, and prediction horizons of $T_o=2$, $T_a=8$, and $T_p=16$, respectively.
Both TacImag and the downstream diffusion policy are trained with 100 diffusion timesteps and use 10-step DDIM~\cite{DDIM} sampling during inference.
Each task contains 50 demonstrations, except bulb installation, which uses 20 demonstrations due to its longer trajectories.
Reported success rates are averaged over the final 10 training checkpoints, 50 randomized environment initializations, and 3 random seeds, resulting in 1500 rollouts per policy.

\subsection{Simulation Results}
\begin{table*}[t]
\centering
\caption{Success rates (\%) in simulation across six contact-rich manipulation tasks.}
\label{tab:sim}

\begin{tabular}{lccccccc}
\toprule
Configuration
& USB Insertion & Power Insertion & Peg-in-hole & Gear Assembly & Bulb Installation & Ball Sorting & Avg. \\
\midrule

Vision
& 52.7 & 52.5 & 24.4 & \textbf{60.7} & 75.2 & 56.1 & 53.6 \\

Vision + TacRGB
& 51.8 & 59.4 & 16.3 & 53.4 & 77.8 & \textbf{63.5} & 53.7 \\

Vision + TacFF
& \textbf{57.5} & \textbf{63.0} & \underline{29.2} & 57.8 & \textbf{82.4} & 59.1 & \textbf{58.2} \\

\midrule

Vision + TacRGB (Imag.)
& 53.3 & 52.3 & 23.8 & 55.1 & 75.0 & \underline{59.7} & 53.2 \\

Vision + TacFF (Imag.)
& \underline{56.2} & \underline{61.2} & \textbf{31.3} & \underline{58.7} & \underline{80.2} & 57.5 & \underline{57.5} \\

\bottomrule
\end{tabular}

\vspace{2mm}

\footnotesize
\parbox{0.93\linewidth}{
Vision uses only visual observations and proprioception.
TacRGB and TacFF denote policies equipped with physical tactile sensing.
Imag. variants replace physical tactile measurements with tactile observations generated by TacImag and therefore require no tactile sensor during deployment.
Results are averaged over 3 random seeds, 50 environment initializations, and the final 10 checkpoints.
}

\end{table*}

\begin{figure}[t]
    \centering
    \includegraphics[width=1.0\columnwidth]{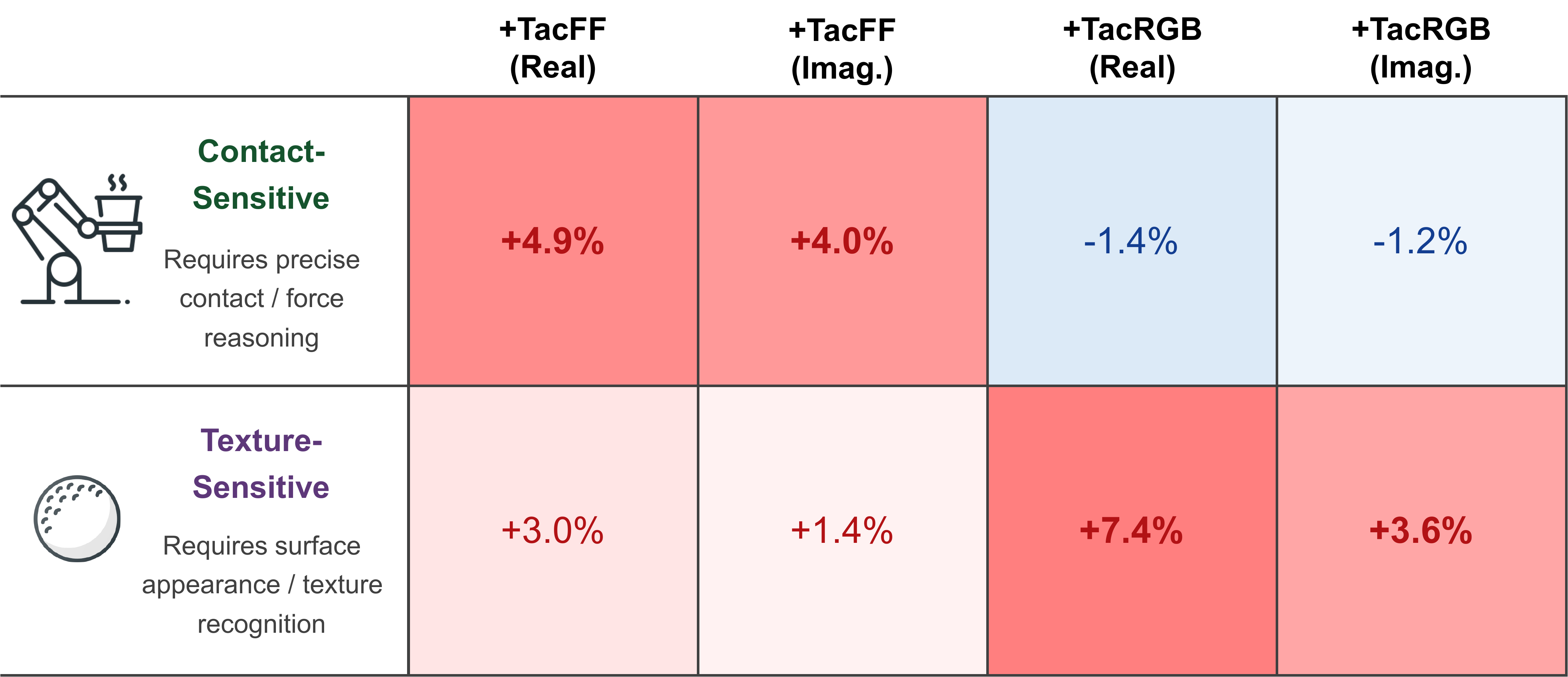}
    \caption{
    \textbf{Average success-rate improvement (\%) over the Vision baseline for contact-sensitive and texture-sensitive tasks.}
    Contact-sensitive tasks correspond to USB insertion, power-plug insertion, peg-in-hole, gear assembly, and bulb installation, while texture-sensitive tasks correspond to ball sorting.
    Physical tactile sensing and imagined tactile observations exhibit consistent task-dependent trends: TacFF provides larger gains on contact-sensitive tasks, whereas TacRGB is more beneficial for texture-sensitive tasks.
    The comparable improvements achieved by imagined tactile observations indicate that TacImag successfully captures task-relevant tactile information without requiring tactile hardware during deployment.
    }
    \label{fig:heatmap}
\end{figure}

Table~\ref{tab:sim} reports success rates across six simulated manipulation tasks.
Among all configurations, physical TacFF achieves the highest average success rate ($58.2\%$), followed closely by imagined TacFF ($57.5\%$), which requires no tactile sensing during deployment.
Compared with the vision-only baseline ($53.6\%$), imagined TacFF improves the average success rate by $3.9$ percentage points and remains within $0.7$ points of the corresponding physical tactile policy.

Across individual tasks, imagined TacFF outperforms the vision-only baseline on five of the six tasks and achieves the best overall result on peg-in-hole ($31.3\%$), slightly exceeding the physical TacFF policy ($29.2\%$).
One possible explanation is that the diffusion model learns a smoother, task-oriented tactile representation that suppresses irrelevant variability while preserving manipulation-relevant contact cues, making it easier for the policy to exploit.
Given the relatively small margin, however, we do not interpret this result as evidence that imagined tactile is inherently superior to physical tactile. For USB insertion, power insertion, gear assembly, and bulb installation, the gap between imagined and physical TacFF remains below $2.2$ percentage points, indicating that TacImag preserves most of the task-relevant tactile information required for contact-rich manipulation.

To better understand the role of tactile imagination, Fig.~\ref{fig:heatmap}
groups tasks according to their dominant tactile requirements.
Contact-sensitive tasks (USB insertion, power insertion, peg-in-hole, gear assembly, and bulb installation) primarily depend on contact geometry and force feedback, whereas the texture-sensitive ball-sorting task relies on surface appearance and texture cues.

Consistent with prior observations in visuotactile manipulation, TacFF provides the largest gains on contact-sensitive tasks, while TacRGB is most beneficial on the texture-sensitive task.
More importantly, imagined tactile observations exhibit the same task-dependent trends as their physical counterparts.
Imagined TacFF achieves a $4.0\%$ average improvement over the vision baseline on contact-sensitive tasks, closely matching the $4.9\%$ improvement obtained by physical TacFF.
Similarly, imagined TacRGB improves performance by $3.6\%$ on the texture-sensitive sorting task, recovering a substantial portion of the gain provided by physical TacRGB ($7.4\%$).
These results suggest that TacImag captures task-relevant tactile information rather than merely generating visually plausible tactile observations.

\subsection{Similarity Analysis}
\begin{table}[t]
\centering
\caption{Tactile imagination quality on held-out validation episodes.}
\label{tab:metric}
\renewcommand{\arraystretch}{1.1}
\begin{tabular}{lccc}
\toprule
Task &
TacFF Cos$\uparrow$ &
TacRGB SSIM$\uparrow$ &
TacRGB LPIPS$\downarrow$ \\
\midrule
USB Insertion     & 0.854 & 0.952 & 0.138 \\
Power Insertion   & 0.963 & 0.972 & 0.107 \\
Peg-in-Hole       & 0.937 & 0.960 & 0.158 \\
Gear Assembly     & 0.897 & 0.920 & 0.178 \\
Bulb Installation & 0.793 & 0.973 & 0.153 \\
Ball Sorting      & 0.896 & 0.942 & 0.207 \\
\midrule
Average           & 0.890 & 0.953 & 0.157 \\
\bottomrule
\end{tabular}

\vspace{1mm}

\footnotesize
\parbox{0.95\linewidth}{
TacFF is evaluated using force-direction cosine similarity.
TacRGB is evaluated using structural similarity (SSIM) and learned perceptual image patch similarity (LPIPS).
Metrics are computed on held-out validation episodes.
}
\end{table}

Table~\ref{tab:metric} quantitatively evaluates the quality of imagined tactile observations on held-out validation episodes.
For TacFF, TacImag achieves an average force-direction cosine similarity of $0.890$ across all tasks, reaching $0.963$ on power insertion and remaining above $0.79$ even for the most challenging bulb-installation task.
For TacRGB, TacImag achieves an average SSIM of $0.953$ and LPIPS of $0.157$, indicating high structural and perceptual similarity to physical tactile images.

Together, these results demonstrate that TacImag can accurately reconstruct both force-field and tactile-image representations from visual observations alone.
Combined with the policy results in Table~\ref{tab:sim} and Fig.~\ref{fig:heatmap}, these findings suggest that the generated tactile observations retain the task-relevant information required for downstream manipulation.

\subsection{Case Studies}
\begin{figure*}[t]
    \centering
    \includegraphics[width=\textwidth]{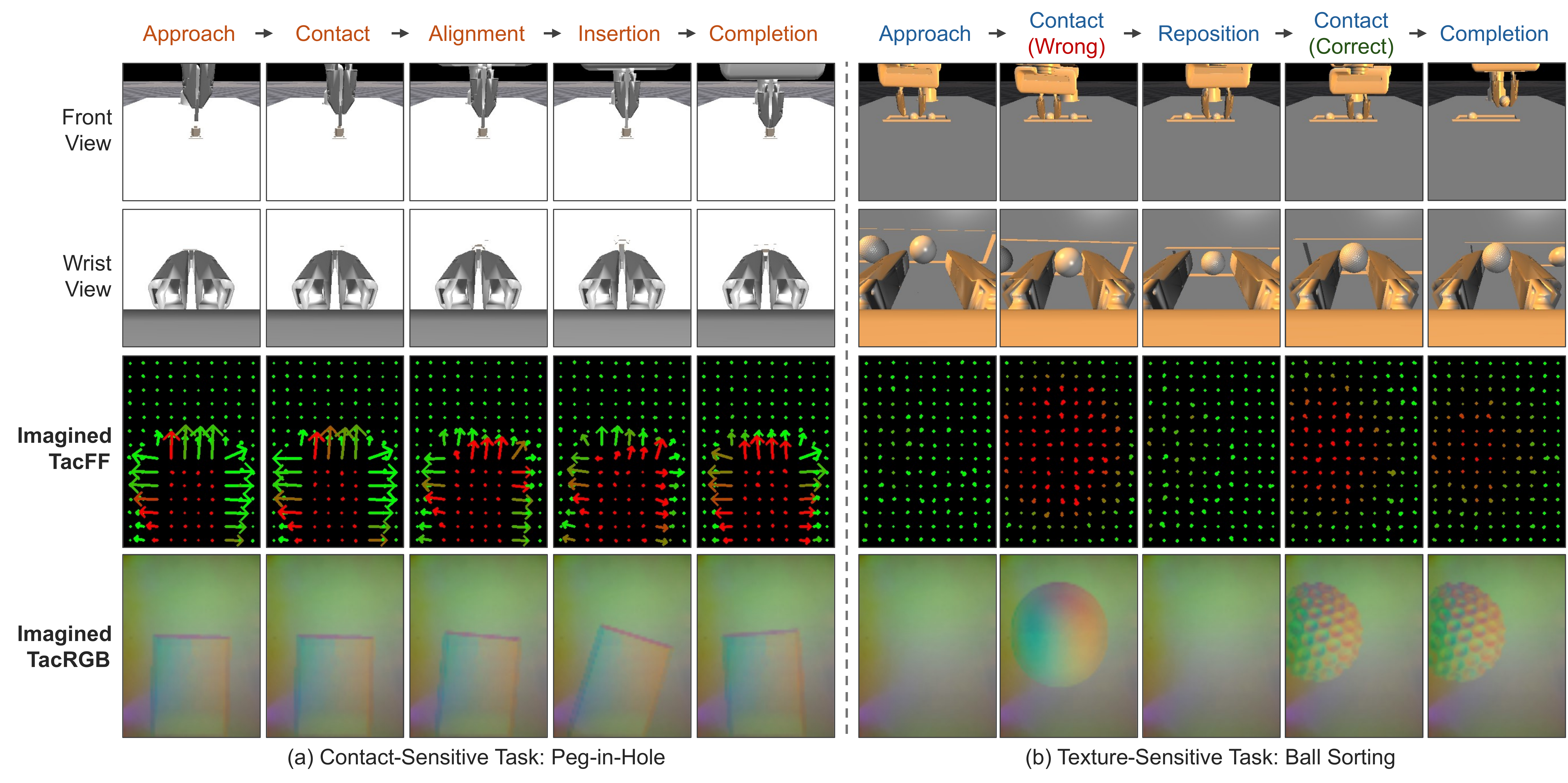}
    \caption{
    \textbf{Case studies of tactile imagination on contact-sensitive and texture-sensitive tasks.}
    \textbf{(a)} In the peg-in-hole task, imagined TacFF remains weak before contact and becomes structured after contact, revealing force directions and contact geometry that facilitate alignment and insertion.
    \textbf{(b)} In the ball sorting task, imagined TacRGB captures discriminative surface appearance and texture cues.
    When the gripper contacts the wrong object, the imagined TacRGB shows a smooth appearance corresponding to the ping-pong ball.
    After repositioning to the target golf ball, the imagined TacRGB reveals the characteristic dimple pattern, enabling successful object identification.
    For visualization, both front and wrist views are shown.
    During tactile imagination, contact-sensitive tasks use the wrist view as input, while the texture-sensitive sorting task uses the front view as input.
    }
    \label{fig:case_study}
\end{figure*}

Figure~\ref{fig:case_study} provides qualitative examples of tactile imagination for both contact-sensitive and texture-sensitive tasks.

For peg-in-hole, imagined TacFF remains largely unstructured before contact and gradually develops coherent force patterns after contact is established.
As insertion progresses, the predicted force field reveals contact geometry and force directions that are difficult to infer from vision alone, providing additional cues for alignment and insertion.

For ball sorting, imagined TacRGB captures discriminative texture information that is not directly observable from the selected camera viewpoint.
When the gripper first contacts a ping-pong ball, the imagined tactile image exhibits a smooth appearance.
After repositioning to the target golf ball, the imagined tactile image reveals the characteristic dimple pattern of the golf-ball surface.
These observations indicate that TacImag can infer task-relevant tactile properties from visual context and use them to support downstream decision making.

Together, these examples illustrate two complementary roles of tactile imagination: predicting contact and force information for contact-rich manipulation, and recovering texture information for object recognition and classification.

\section{Real-World Experiments}

\subsection{Real-World Setup}
\begin{figure*}[t]
    \centering
    \includegraphics[width=0.9\textwidth]{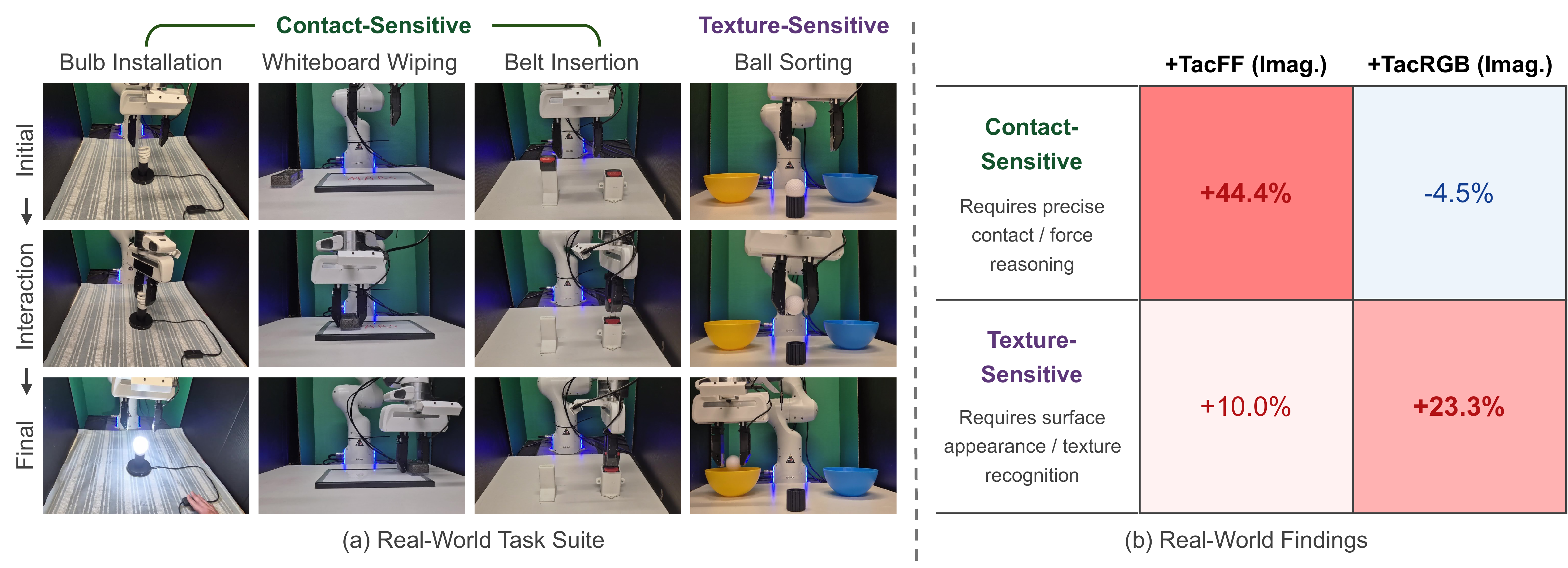}
    \caption{
    \textbf{Real-world experiments and findings.}
    \textbf{(a) Real-World Task Suite.}
    Real-world manipulation tasks including bulb installation, whiteboard wiping, belt insertion, and ball sorting.
    Each task is visualized at three representative stages: initial, interaction, and final.
    \textbf{(b) Real-World Findings.}
    Average success-rate improvement (\%) over the vision-only baseline.
    Contact-sensitive tasks correspond to bulb installation, whiteboard wiping, and belt insertion, while texture-sensitive tasks correspond to ball sorting.
    Imagined TacFF provides the largest gains on contact-sensitive tasks, whereas imagined TacRGB is most effective for texture-sensitive object discrimination.
    Results for ball sorting are averaged over front-view and wrist-view settings.
    }
    \label{fig:real_summary}
\end{figure*}
The real-world setup is shown in Fig.~\ref{fig:real_summary}(a).
Experiments are conducted on a Franka Emika Panda robot arm equipped with a parallel-jaw gripper, a wrist-mounted Intel RealSense D415 camera, and a front-facing Intel RealSense D415 camera.
During data collection, a GelSight R1.5 sensor~\cite{wang2021gelsight} is mounted on the right gripper finger to record paired visuotactile demonstrations for training the TacImag model.
After training, the sensor is removed and replaced with a dummy finger sharing the same external geometry.
As a result, all deployment experiments are performed without tactile sensing, while tactile observations are generated online by the frozen TacImag model using only visual observations and proprioceptive states.

We evaluate TacImag on four real-world manipulation tasks: bulb installation, whiteboard wiping, belt insertion, and ball sorting.
Bulb installation, whiteboard wiping, and belt insertion are contact-sensitive tasks that require accurate force reasoning during interaction.
Ball sorting is a texture-sensitive task that requires recognizing surface appearance cues to distinguish golf balls from ping-pong balls.

For bulb installation, the bulb is initially inserted into the socket with a randomized amount of looseness.
The robot must rotate the bulb until sufficient contact is established to complete the electrical circuit and illuminate the light.
For whiteboard wiping, both the whiteboard and eraser are fixed, while the written content varies in size, orientation, and location.
Successful wiping requires maintaining consistent contact between the eraser and the board surface.
For belt insertion, both the belt and insertion slot remain fixed, but successful completion requires applying sufficient insertion force while preventing slippage.
For ball sorting, the bowl locations remain fixed while the object identity varies.
The robot must place golf balls into the yellow bowl and ping-pong balls into the blue bowl.

All policies are trained using the same diffusion-policy architecture as in simulation.
Each configuration is evaluated over 15 real-world trials.
Following our simulation findings, contact-sensitive tasks use wrist-view observations to generate imagined TacFF, while ball sorting is evaluated using both wrist-view and front-view observations to study the effect of viewpoint on tactile imagination.

\subsection{Real-World Results}
\begin{table*}[t]
\centering
\caption{Success rates (\%) in real-world contact-rich manipulation tasks.}
\label{tab:real}

\renewcommand{\arraystretch}{1.15}

\begin{tabular}{lccccc}
\toprule
Configuration
&
\begin{tabular}[c]{@{}c@{}}
Bulb Installation\\
(Wrist View)
\end{tabular}
&
\begin{tabular}[c]{@{}c@{}}
Whiteboard Wiping\\
(Wrist View)
\end{tabular}
&
\begin{tabular}[c]{@{}c@{}}
Belt Insertion\\
(Wrist View)
\end{tabular}
&
\begin{tabular}[c]{@{}c@{}}
Ball Sorting\\
(Front View)
\end{tabular}
&
\begin{tabular}[c]{@{}c@{}}
Ball Sorting\\
(Wrist View)
\end{tabular}
\\
\midrule

Vision
& 26.7 & 20.0 & 6.7 & 26.7 & 33.3 \\

Vision + TacRGB (Imag.)
& 33.3 & 6.7 & 0.0 & \textbf{33.3} & \textbf{73.3} \\

Vision + TacFF (Imag.)
& \textbf{86.7} & \textbf{60.0} & \textbf{40.0} & \textbf{33.3} & \underline{46.7} \\

\bottomrule
\end{tabular}

\vspace{1mm}

\footnotesize
\parbox{0.8\linewidth}{
Results are reported as success rates (\%).
All tactile observations are generated by TacImag and no tactile sensor is used during deployment.
Each configuration is evaluated over 15 real-world trials.
Tasks marked as \emph{Wrist View} use wrist-camera observations for tactile imagination, while \emph{Front View} uses front-camera observations.
}
\end{table*}

Table~\ref{tab:real} summarizes the real-world results, while Fig.~\ref{fig:real_summary}(b) groups tasks according to their tactile requirements.

\textbf{Imagined TacFF substantially improves contact-sensitive manipulation.}
Across bulb installation, whiteboard wiping, and belt insertion, imagined TacFF consistently provides the largest performance gains.
Compared with the vision-only baseline, success rates improve from 26.7\% to 86.7\% on bulb installation, from 20.0\% to 60.0\% on whiteboard wiping, and from 6.7\% to 40.0\% on belt insertion.
Averaged across these three tasks, imagined TacFF improves success rates by 44.4 percentage points over vision-only policies.
These results demonstrate that TacImag can successfully transfer contact-related tactile information into real-world manipulation without requiring tactile hardware during deployment.

\textbf{Imagined TacRGB benefits texture-sensitive object discrimination.}
On ball sorting, imagined TacRGB achieves the highest performance among all evaluated configurations.
Using wrist-view observations, success rates increase from 33.3\% to 73.3\%, substantially outperforming imagined TacFF (46.7\%).
This result mirrors the trends observed in simulation, where texture-sensitive tasks benefit more from tactile appearance representations than force-field representations.

\textbf{Task-dependent trends transfer from simulation to reality.}
The same relationship between tactile representation and task characteristics observed in simulation also appears in the real world.
Imagined TacFF is most effective for contact-sensitive tasks requiring force reasoning, whereas imagined TacRGB is most effective for texture-sensitive object discrimination.
These findings suggest that TacImag preserves task-relevant tactile information despite relying solely on visual observations at deployment.

\subsection{Discussion}
\begin{figure*}[t]
    \centering
    \includegraphics[width=0.85\textwidth]{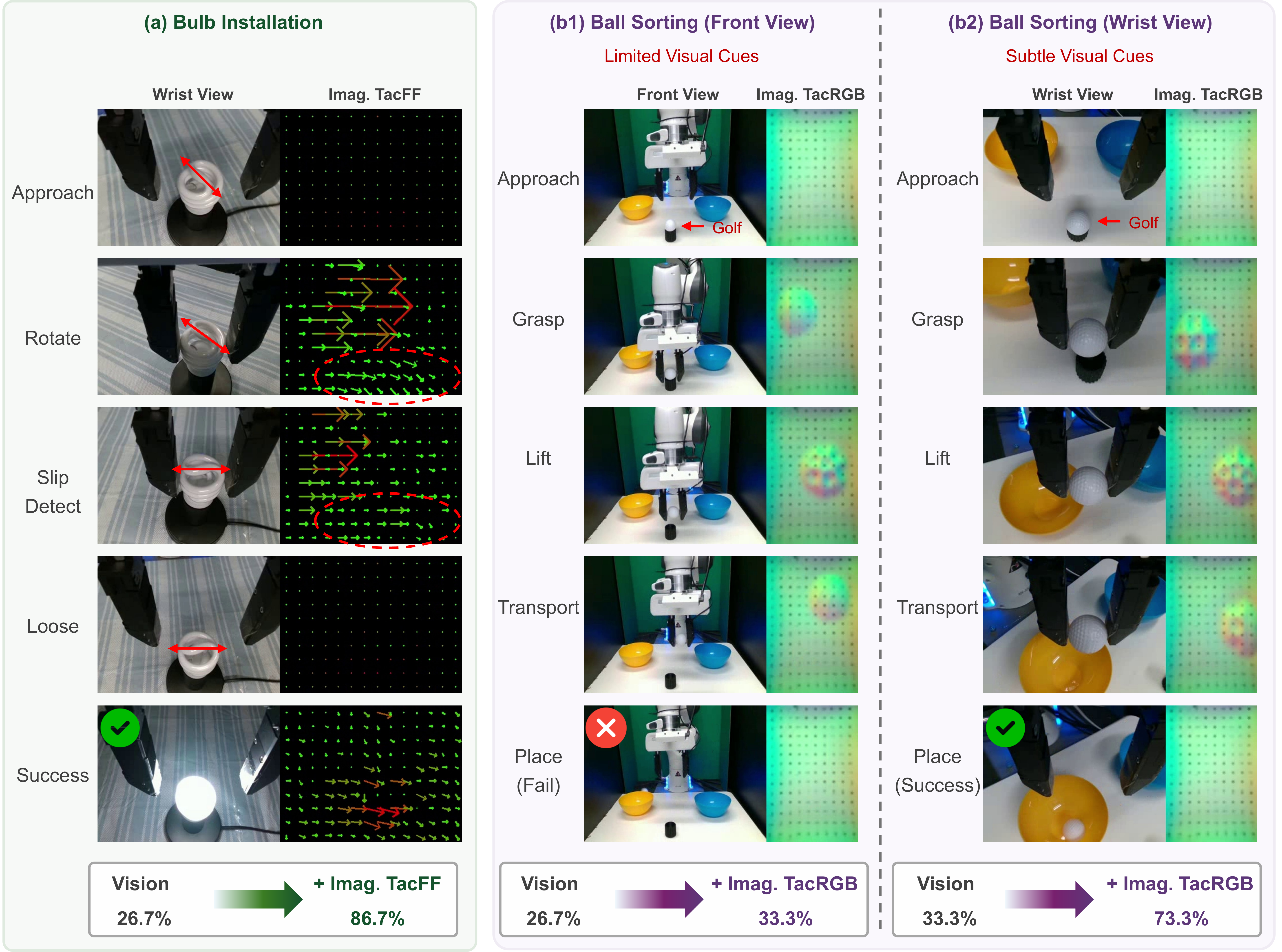}
    \caption{
    \textbf{Real-world tactile imagination rollouts.}
    \textbf{(a)} In bulb installation, imagined TacFF becomes structured after contact and revealing force directions associated with alignment, rotation, and localized slip events, leading to a large improvement over the vision-only policy ($26.7\% \rightarrow 86.7\%$).
    \textbf{(b1--b2)} In ball sorting, TacImag exhibits viewpoint-dependent behavior.
    With front-view observations, the visual cues for surface texture are weak, resulting in only limited improvement from imagined TacRGB ($26.7\% \rightarrow 33.3\%$).
    With wrist-view observations, local surface cues become more visible during grasping and transport, allowing imagined TacRGB to reveal discriminative texture patterns and substantially improve performance ($33.3\% \rightarrow 73.3\%$).
    These results suggest that TacImag does not hallucinate new information, but instead converts subtle visual cues into tactile representations that are easier for the policy to exploit.
    }
    \label{fig:real_rollout}
\end{figure*}

Figure~\ref{fig:real_rollout} provides qualitative examples of tactile imagination in real-world manipulation.

\textbf{TacImag generates meaningful force patterns during contact-rich manipulation.}
In bulb installation, imagined TacFF remains largely inactive before contact and becomes increasingly structured once contact is established.
During grasping, rotation, and tightening, the generated force field reveals directional force patterns consistent with the interaction between the gripper, bulb, and socket.
Notably, the imagined TacFF also highlights localized shear patterns associated with sliding contact during bulb rotation, suggesting that TacImag captures interaction dynamics beyond simple contact detection.
These force cues provide useful information for detecting contact, maintaining grip stability, and preventing slippage, which helps explain the substantial performance improvements observed in Table~\ref{tab:real}.

\textbf{TacImag is constrained by the information available in vision.}
The ball-sorting task reveals an important limitation and insight of tactile imagination.
When only front-view observations are provided, the visual cues distinguishing golf balls from ping-pong balls are weak.
Consequently, the generated TacRGB observations remain ambiguous and provide only limited performance gains (26.7\% to 33.3\%).
In contrast, wrist-view observations expose subtle surface details during grasping and transport.
TacImag transforms these subtle visual cues into explicit tactile appearance patterns, enabling substantially better object discrimination and increasing success rates from 33.3\% to 73.3\%.

Together, these observations suggest that TacImag does not hallucinate tactile information absent from the visual observations.
Instead, it acts as a task-dependent representation transformation that converts visually observable but difficult-to-use interaction cues into tactile representations that are easier for policies to exploit.
This perspective explains why tactile imagination can improve manipulation performance without requiring physical tactile sensing during deployment.

%%%%%%%%%%%%%%%%%%%%%%%%%%%%%%%%%%%%%%%%%%%%%%%%%%%%%%%%%%%%%%%%%%%%%%%%%%%%%%%%
\section{CONCLUSIONS}
We presented \textit{TacImag}, a tactile imagination framework that enables touch-informed manipulation without requiring tactile sensors during deployment. By learning a vision-to-touch diffusion model from paired visuotactile demonstrations, TacImag generates tactile representations online and provides them to a tactile-conditioned manipulation policy.

Across six simulation tasks and four real-world tasks, TacImag consistently improved manipulation performance using only visual observations and proprioceptive states at execution time. We found that the effectiveness of tactile imagination depends strongly on both tactile representation and task characteristics: imagined force fields are most beneficial for contact-sensitive manipulation, whereas imagined tactile images provide larger gains for texture-sensitive object discrimination. These trends closely mirror those observed with physical tactile sensing.

Our results further suggest that tactile imagination acts as a form of contact-aware supervision rather than hallucinating information absent from vision. By transforming subtle visual interaction cues into tactile representations, TacImag enables policies to better exploit contact-relevant information while avoiding the complexity of tactile hardware. Future work will investigate long-horizon manipulation, richer tactile representations, and the integration of tactile imagination with world models and active perception.

%%%%%%%%%%%%%%%%%%%%%%%%%%%%%%%%%%%%%%%%%%%%%%%%%%%%%%%%%%%%%%%%%%%%%%%%%%%%%%%%

\bibliographystyle{unsrt}
\bibliography{references}

\end{document}